\documentclass[
]{ceurart}

\usepackage{listings}
\usepackage{cleveref}

\usepackage{csquotes}
\usepackage{float}
\usepackage{ragged2e}
\usepackage{lipsum}

\usepackage{tikz}
\usetikzlibrary{decorations.pathmorphing}
\usetikzlibrary{decorations.pathreplacing}
\usetikzlibrary{intersections,backgrounds}
\usetikzlibrary{calc}

\usepackage[nolist]{acronym}

\newcolumntype{x}[1]{>{\centering\let\newline\\\arraybackslash\hspace{0pt}}p{#1}}
\newcolumntype{y}[1]{>{\raggedright\let\newline\\\arraybackslash\hspace{0pt}}p{#1}}

\begin{document}

\ifcasreviewlayout
    \linenumbers
\else
\fi

\copyrightyear{2024}
\copyrightclause{Copyright for this paper by its authors. Use permitted under Creative Commons License Attribution 4.0 International (CC BY 4.0).}

\conference{EWAF'24: European Workshop on Algorithmic Fairness, July 01--03, 2024, Mainz, Germany}

\title{Mapping the Potential of Explainable AI for Fairness Along the AI Lifecycle}

\author[1,2]{Luca Deck}[email=luca.deck@uni-bayreuth.de]
\author[1]{Astrid Schomäcker}[email=astrid.schomaecker@uni-bayreuth.de]
\author[1]{Timo Speith}[email=timo.speith@uni-bayreuth.de]
\author[3]{Jakob Schöffer}[email=schoeffer@utexas.edu]
\author[1]{Lena Kästner}[email=lena.kaestner@uni-bayreuth.de]
\author[1,2]{Niklas Kühl}[email=kuehl@uni-bayreuth.de]

\address[1]{University of Bayreuth, Germany}
\address[2]{Fraunhofer FIT, Germany}
\address[3]{University of Texas at Austin, USA}



\begin{abstract}
The widespread use of artificial intelligence (AI) systems across various domains is increasingly surfacing issues related to algorithmic fairness, especially in high-stakes scenarios.
Thus, critical considerations of how fairness in AI systems might be improved---and what measures are available to aid this process---are overdue.
Many researchers and policymakers see explainable AI (XAI) as a promising way to increase fairness in AI systems. 
However, there is a wide variety of XAI methods and fairness conceptions expressing different desiderata, and the precise connections between XAI and fairness remain largely nebulous. 
Besides, different measures to increase algorithmic fairness might be applicable at different points throughout an AI system's lifecycle.
Yet, there currently is no coherent mapping of fairness desiderata along the AI lifecycle. 
In this paper, we we distill eight fairness desiderata, map them along the AI lifecycle, and discuss how XAI could help address each of them.
We hope to provide orientation for practical applications and to inspire XAI research specifically focused on these fairness desiderata.
\end{abstract}


\maketitle

\section{Introduction}

The emergence and widespread use of artificial intelligence (AI) systems across various sectors and domains is increasingly shifting attention from considerations of mere performance to considerations about algorithmic fairness. 
This is particularly relevant for systems employed in high-stakes scenarios and especially pressing in contexts prone to harmful societal biases \citep{Barocas.2016, Ntoutsi.2020}.
This has sparked a growing demand for approaches to scrutinize and improve fairness in AI systems.
In the literature, there is a common recognition that fairness in AI systems demands various perspectives and measures (e.g., \citep{DeArteaga.2022, Starke.2022, AlerTubella.2023}).
In this paper, we set out to integrate two strategies that have been suggested:
First, a growing community of researchers proposes explainable AI (XAI) as a versatile and powerful tool to combat unfairness \citep{Deck.ACritical.2024}. 
Second, others have focused on the AI lifecycle and tried to determine where fairness issues originate (e.g., \cite{fazelpour2021algorithmic, agarwal2023seven}). 
They hope that once identified, fairness issues can be mitigated by taking appropriate steps in the relevant phase within the lifecycle. 
 
While both these strategies seem intuitively promising, neither currently presents a satisfactory and comprehensive picture. 
A primary reason why neither approach currently fulfills their potential is, we take it, that there are various types and kinds of fairness discussed in the literature, yielding different \emph{fairness desiderata}. 
Differentiating between these desiderata is crucial to gain an overall picture of which measures are most promising to address fairness in which contexts.
Without such differentiation, the utility of XAI is not as straightforward as commonly claimed. Accordingly, there is a need for more clarity about how, exactly, XAI contributes to which fairness desideratum \citep{Deck.ACritical.2024}.
Existing discussions on this matter suffer, by and large, from both a too narrow conception of XAI and a mostly under-specified notion of fairness. 
Similarly, existing attempts to map measures for achieving fairness onto the AI lifecycle remain crucially incomplete, for they have limited their attention to only a subset of the relevant fairness desiderata.
To overcome these limitations, we distill eight fairness desiderata, map them along the AI lifecycle, and discuss how XAI can help address each of them.

For the purposes of this paper, we will focus specifically on how XAI can be utilized to improve AI systems' fairness throughout their lifecycle.
Concerning this question, we aim to develop a holistic account based on the fairness desiderata we distill.
While our proposal is not intended as a guideline, we believe it will stimulate scientific discussions about and practical applications of fairness measures and, as such, will be a valuable starting point to explore fairness opportunities for researchers, developers, and regulators alike.

We begin by introducing preliminaries and diagnosing the problem in \Cref{sec:background}. 
Importantly, we do not limit our understanding of algorithmic fairness to computational perspectives. In \Cref{sec:mapping} we propose eight fairness desiderata from interdisciplinary literature: fairness understanding, data fairness, formal fairness, perceived fairness, fairness with human oversight, empowering fairness, long-term fairness, and informational fairness (contribution \#1).
We map our fairness desiderata along the different stages within the AI lifecycle and suggest that each fairness desideratum affords a different entry point for taking measures to improve fairness.
This closes the first gap in the current literature as it highlights where in the lifecycle different fairness desiderata become especially relevant (contribution \#2). 
Utilizing this mapping, we discuss how XAI can be leveraged to address the different fairness desiderata at the respective points throughout the AI lifecycle.
This closes the second gap in the current literature as it allows us to systematically examine the potential of XAI to address algorithmic fairness in different circumstances (contribution \#3). 
To illustrate the utility of our approach, we discuss its application to the COMPAS case (see \cite{Angwin.2022}) in \Cref{sec:compas}. 
Before closing, we point to some avenues for future research in \Cref{sec:conclusion}.

\section{Background}\label{sec:background}

In this article, we contribute to closing two research gaps: 1) considerations of fairness along the AI lifecycle are incomplete, and 2) it is unclear how exactly XAI can help foster fairness.
Before closing these gaps, we retrace contributions and debates of prior works.

\subsection{Algorithmic Fairness}
\label{AIfairness}

The debate on algorithmic fairness draws inspiration from various disciplines.
Several scandals surrounding AI systems that disadvantage marginalized groups \cite{Angwin.2022, Dastin.2022, FUSTER.2022} have shattered early hopes put into the \enquote{neutrality} of AI \citep{Barocas.2016}.
In reaction, scholars from computer science, philosophy, social science, law, and psychology have been engaging in debates on what it means for algorithmic decision-making to be fair (see \cite{Mulligan.2019} for interdisciplinary perspectives).

The technical debate on algorithmic fairness primarily focuses on formal measures of fairness \cite{Barocas.2019, Verma.2018, Castelnovo.2022}.
After it has proven unhelpful to remove information about membership in marginalized groups from training data (``fairness through unawareness'' \cite{Dwork.2012}), most approaches from the field of computer science rely on comparing the outcomes for people from different groups. 
An important finding is that the different formal measures for fairness can, in most cases, not be fulfilled simultaneously \cite{Kleinberg.2017, Defrance.2023}.
Thus, (formally) fair AI systems require crucial design choices about which formal measures to apply in which context.

To aid such decisions, philosophers have connected formal measures to different philosophical theories \cite{Binns.2018, lee_M.S.A.2021, Khan.2022}. 
In general, \enquote{fairness} is a normative concept---and debates about fairness can be understood as a way to discuss what is morally right or wrong in a given case (see \cite{Mulligan.2019}).
More specifically, fairness can be understood as an issue of justice, non-discrimination, or equality \cite{Binns.2018}, which connects it to numerous strands of philosophical discussion. 

Beyond that, social scientists have pointed out shortcomings of existing formal methods, e.g., problems in detecting issues of intersectionality \cite{crenshaw_mapping_1991} and their reliance on constructed categories like race and gender \cite{hanna_towards_2020}.
Legal scholars are primarily interested in how legislation on discrimination can be applied to algorithmic decision-making and whether additional regulation is required \cite{Wachter.2021, Barocas.2016}.
Psychological research is generally interested in whether algorithmic decision-making is perceived as fair \citep{Starke.2022,Lee.2018}.
For example, \citet{Colquitt.2015} distinguish four different psychological dimensions of fairness: Whether the outcome is fair (viz., distributive justice), whether the process that leads to the outcome is fair (viz., procedural justice), whether the information about the decision is communicated truthfully and thoroughly (viz., informational justice), and whether individuals are treated respectfully (viz., interpersonal fairness). 

Overall, algorithmic fairness can be addressed from several different angles---some of which are complementary, some of which are contradictory \citep{Friedler.2021, Defrance.2023}.
This stresses the need to be clear about the specific desideratum behind any attempt to improve fairness.  

\subsection{Considerations of Fairness Along the AI Lifecycle Remain Incomplete}
\begin{figure}[htbp]
    \centering
    \includegraphics[width=\textwidth]{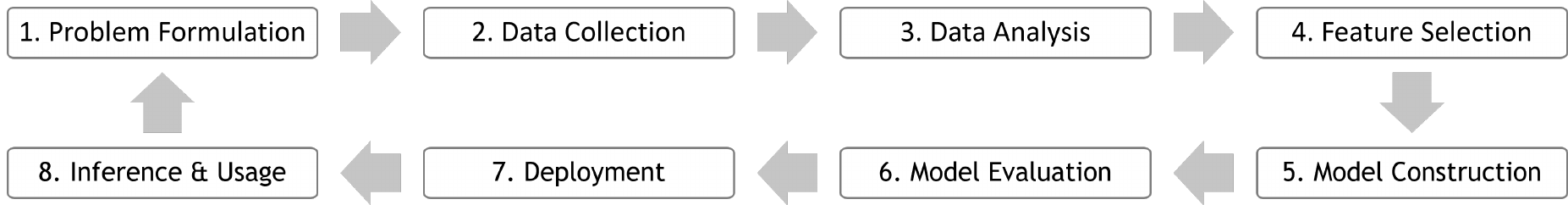}
    \caption{The AI Lifecycle we use, combined from \citeauthor{quemy2020two}~\cite{quemy2020two} and \citeauthor{wang2017machine}~\cite{wang2017machine}.}
    \label{fig:lifecycle}
\end{figure}

A common strategy to determine potential entry points to address fairness issues involves examining each step of the AI lifecycle. 
There have been several proposals of prototypical AI lifecycles (e.g., \cite{Wirth2000CrispdmTA, fayyad1996data, wang2017machine, quemy2020two, kuhl2021conduct}), each with slightly different stages or adapted to different application areas.
The AI lifecycle used in this article is a combination of the proposals by \citet{quemy2020two} and \citet{wang2017machine} involving the following stages: 1) problem formulation, 2) data collection, 3) data analysis, 4) feature selection, 5) model construction, 6) model evaluation, 7) deployment, and 8) inference and usage (see \Cref{fig:lifecycle}). 

First, \emph{problem formulation} involves abstraction and formulation of the problem, which is to be solved using AI.
Afterward, \emph{data collection} aims to establish a representative data set suitable to develop an AI system able to solve the defined problem. 
\emph{Data analysis} includes descriptive statistics to understand the characteristics of the data at hand and data pre-processing to prepare the data for further operations.
Subsequently, in the \emph{feature selection} stage, features are excluded, transformed, or aggregated for effective and efficient handling in the training process. 
\emph{Model construction}, then, includes the selection of the training algorithm and the training itself.
Iteratively, performance objectives are tested during \emph{model evaluation} and optimized through algorithm adjustments, e.g., parameter tuning.  
\emph{Deployment} refers to the integration of the AI system into a productive environment once the model achieves sufficient performance.
Finally, \emph{inference \& usage} describes concrete output generated by the AI system and its impact on the surroundings, such as the business context, society, and environment.
Note that these stages are often not entered sequentially, leaving room for iterations and loops between stages \citep{Kreuzberger2023MLOPS}.

Much research has focused on what types of bias can emerge in or affect different steps of the AI lifecycle (e.g., \citep{DeArteaga.2022, agarwal2023seven, lee2021risk, singh2022fair}). For example,
\citet{singh2022fair} take the CRISP-DM model \citep{Wirth2000CrispdmTA} and review the types of bias that can occur at different stages of this model.
However, these considerations remain incomplete, as most of these works have a strong technical focus and deal only with formal notions of fairness, e.g., by providing guidance for choosing fairness metrics or debating the role of potential fairness-utility tradeoffs (e.g., \citet{Castelnovo.BeFair:Addressing.2020, DeArteaga.2022}).
Against this background, it is important to develop a view that also maps other conceptions of fairness into the lifecycle.

\subsection{Considerations of XAI for Fairness Remain Incomplete} 


To amend the lack of understanding of AI-based systems, their reasoning processes, and their outputs, the research field of XAI emerged in recent years \cite{BarredoArrieta.2020, Paez2019Pragmatic, Chazette2021Exploring, Langer.2021b, Koehl2019Explainability}.
Generally, XAI aims to provide a stakeholder with information about some aspect of an AI system to facilitate their understanding of this aspect \cite{Chazette2021Exploring}. 
Understanding (some aspect of) a system is often just an intermediary step to other goals, such as fairness or appropriate reliance \citep{Langer.2021,schoeffer2024explanations}.
For example, by understanding how a particular system's output came to be, prior work has argued that a person should be enabled to assess whether this output was based on valid criteria or not \cite{Paez2019Pragmatic, Gilpin2019Explaining}.
If an unfavorable decision was based on (a proxy for) the skin color of a person, this person should supposedly be able to recognize that they were treated unfairly \cite{Carvalho2019Machine, Langer2021Auditing}. 

However, the goal of XAI is often narrowly construed as the development of methods to create interpretable surrogate models of black box models~\citep{Rudin.2019, dwivedi2023explainable}.
Such a narrow view of XAI artificially 
restricts the space of options that can be chosen to facilitate stakeholders' understanding of AI systems.
Take, e.g., model cards for model reporting \cite{Mitchell.2019} or datasheets for data sets \cite{Gebru.2021}. 
Although these approaches provide important information about certain aspects of an AI model\footnote{Throughout this paper, we use the term ``AI models'' synonymously for ``machine learning model'' or ``ML model''. While we acknowledge the technical distinction between AI and Machine Learning (ML) as discussed in \citep{kuhl2022artificial}, we adopt the use of the broader ``AI'' term as the prevalent terminology established in the scientific community. This choice reflects the contemporary linguistic trend rather than a lack of distinction between the two phenomena.} (e.g., which training data was used) and thus improve stakeholders' understanding of these aspects, they would rarely be counted as XAI.
Further, XAI methods alone do not provide other kinds of highly relevant information, such as the normative motivations guiding the development of an AI model \citep{Loi.2021}, the analysis of the social context in which it is deployed \citep{Green.2022}, or the descriptive outcome statistics of the deployed model \citep{Chouldechova.2017}.

In a similar vein, the high hopes placed in XAI to mitigate issues of fairness often remain vague or unfulfilled \citep{Deck.ACritical.2024}.
Prior works have criticized the explanatory value \citep{Rudin.2019, Herman.2017}, susceptibility to manipulations \citep{Aivodji.2019, Lakkaraju.2020}, and unsatisfactory interpretations \citep{Balkir.2022, Deck.ACritical.2024} of XAI methods. For example, popular feature-based explanations like LIME~\citep{Ribeiro.2016} or SHAP~\citep{Lundberg.2017} highlighting the use of sensitive features (e.g., gender and race) provide little information about fairness.
This is because these features are often correlated with proxy variables and embedded in use case-specific normative contexts~\citep{Dwork.2012, Lipton.2018b, Nyarko.2021}.

Instead, broadening the conception of XAI enables a more holistic and meaningful mapping of how information about an AI system can contribute to various fairness desiderata.
Against this background, we propose to distinguish the \emph{narrow} conception of XAI, solely pertaining to the inner workings of an AI system, from a \emph{broad} conception of XAI that provides information beyond these inner workings and includes explanations of the broader socio-technical system \citep{dhanorkar2021needs}.
Specifically, we suggest that the broad conception of XAI includes all types of information that increases stakeholders' understanding of (aspects of) an AI system.
For the remainder of this paper, if not indicated otherwise, we refer to the broad conception when we mention XAI.

\section{How XAI Can Be Leveraged for Fairness Along the AI Lifecycle} \label{sec:mapping}

Grounded in a broad interdisciplinary literature review (summarized in \Cref{tab:fairness}), we distill eight categories of what previous work on fairness has called for.
We call these categories \emph{fairness desiderata}, and each desideratum can be instantiated with specific \emph{fairness objectives}.
For the purposes of this paper, we focus specifically on fairness desiderata connected to XAI.
While various stakeholders are involved in utilizing XAI for fairness, an extensive discussion of their various roles is beyond the scope of this paper.
When we mention stakeholders, we rely on the taxonomy provided by \citet{Langer.2021b}.

\begin{figure}
    \centering
    \includegraphics[width=\textwidth]{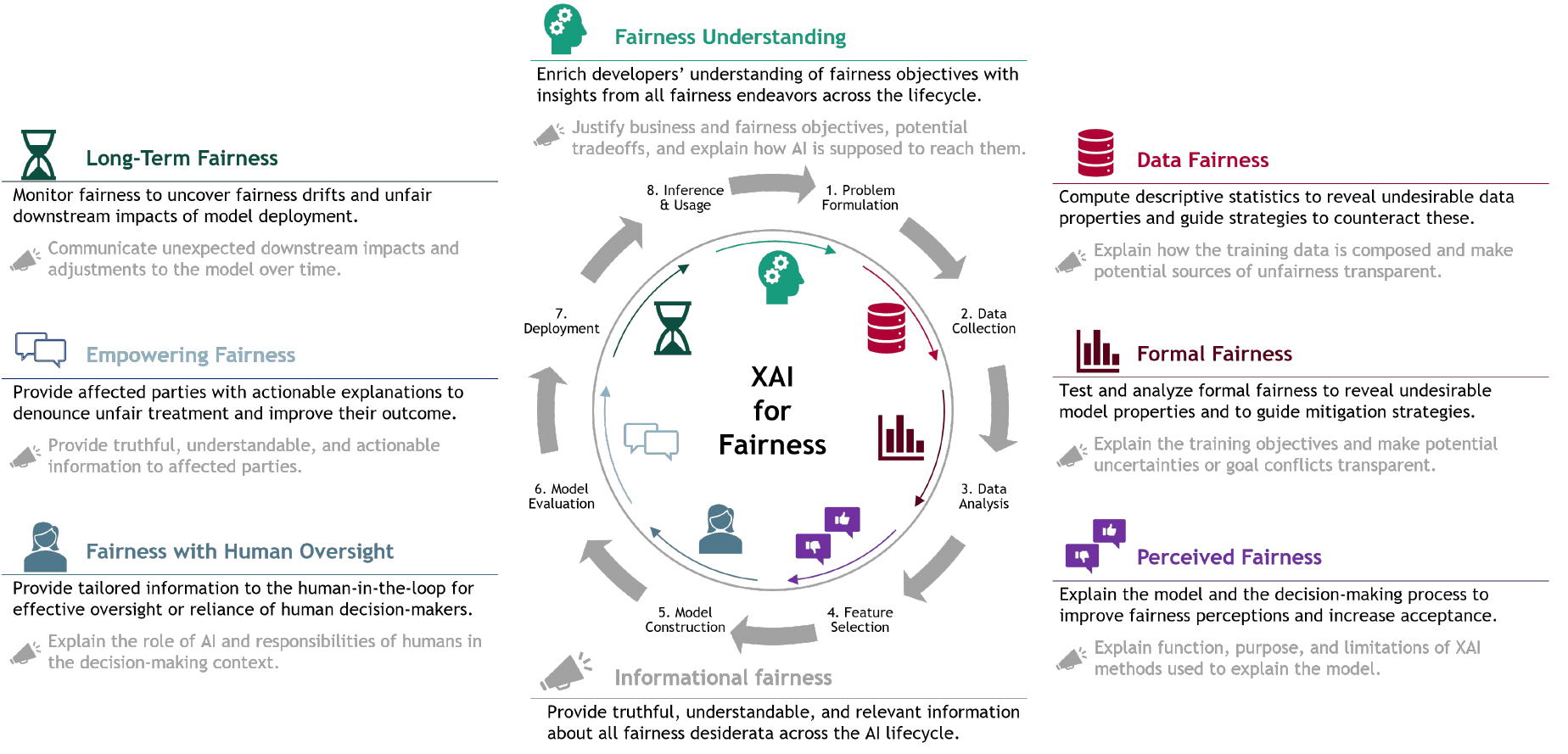}
    \caption{Fairness desiderata along the AI lifecycle depicting how XAI may directly contribute to these desiderata and how it may contribute to informational fairness across all desiderata (symbolized by speakerphones).}
    \label{fig:fairness_lifecycle}
\end{figure}

In what follows, we introduce our eight fairness desiderata, describe the underlying core ideas for each, discuss how they relate to similar concepts, map them onto the AI lifecycle, and elaborate on how we think XAI could contribute to their satisfaction (see \Cref{fig:fairness_lifecycle}).
Note that our proposal does not necessarily present a comprehensive list (see \Cref{sec:conclusion}).

\begin{table}[htbp]
    \caption{Fairness desiderata and their related concepts in existing interdisciplinary literature.} \label{tab:fairness}
    \scriptsize
    \begin{tabular}{p{\dimexpr 0.5\linewidth-2\tabcolsep} |
                   p{\dimexpr 0.5\linewidth-2\tabcolsep}}
    \toprule
    \multicolumn{1}{l}{Fairness desiderata} & Related concepts\\
    \midrule
        \textbf{Fairness understanding} \newline
        Gaining higher-level insights on fairness and the socio-technical challenges surrounding the development and deployment of an AI application to specify concrete fairness objectives.
        & Understanding bias \citep{Ntoutsi.2020}
        \newline Interdisciplinary fairness conceptualization \citep{Mulligan.2019}
        \newline Lessons from political philosophy \citep{Binns.2018}
        \newline Socio-technical perspective \citep{Sartori.2022}\\
    \midrule
        \textbf{Data fairness}\newline
        Identifying and addressing flaws in the data set that might be unfair themselves or potentially lead to downstream violations of fairness objectives.
        & Sampling bias \citep{Mehrabi.2021, DeArteaga.2022} 
        \newline Data errors and bias \citep{Pradhan.2022}
        \newline Data-centric factors in algorithmic fairness \citep{Li.2022}\\
    \midrule
        \textbf{Formal fairness}\newline
        Identifying and addressing model properties leading to violations of formal fairness objectives.
        & Algorithmic bias \citep{DeArteaga.2022}
        \newline Formal fairness definitions \citep{CorbettDavies.2018}
        \newline Fairness metrics \citep{Castelnovo.2022}
        \newline Statistical fairness criteria \citep{Barocas.2019}
        \newline Disparate impact \& disparate treatment \citep{Barocas.2016}\\
    \midrule
        \textbf{Perceived fairness}\newline
        Providing affected parties with explanations and justifications to improve or ``calibrate'' fairness perceptions.
        & Fairness perceptions \citep{Starke.2022, Schoeffer.2022c}
        \newline Perceptions of justice \citep{Colquitt.2001, Binns.2018b}
        \newline Fairness judgment \citep{Dodge.2019}
        \newline Society-in-the loop \citep{Rahwan.2018}\\
    \midrule
        \textbf{Fairness with human oversight}\newline
        Supporting human decision-ma\-kers interacting with an AI system to effectively align human discretion with fairness objectives.
        & Human-ML augmentation for fairness \citep{Teodorescu.2021}
        \newline Appropriate reliance \citep{schemmer2023appropriate}
        \newline Human-in-the-loop \citep{Dodge.2019}
        \newline Domain expertise \citep{Chakraborty.2020}\\
    \midrule
        \textbf{Empowering fairness}\newline
        Providing affected parties with practical information to foster contestability and recourse.
        & Self-informed advocacy \citep{Vredenburgh.2022}
        \newline Procedural fairness \citep{EuropeanCommission.HLEG.2019}
        \newline Counterfactual explanations \citep{Wachter.2017}
        \newline Fair and adequate explanations \citep{Asher.2022}
        \newline The explanation game \citep{Watson.2021}\\
    \midrule
    \textbf{Long-term fairness}\newline
        Monitoring and analyzing the socio-technical long-term impacts of algorithmic decision-ma\-king to adjust unfair repercussions over time.
        & Long-term effects of algorithmic fairness \citep{DeArteaga.2022}
        \newline Fairness drift \citep{Ghosh.2022b}
        \newline Fairness through time \citep{Castelnovo.2021}
        \newline Fairness monitoring \citep{Hardt.2021, Panigutti.2021}\\
    \midrule
    \textbf{Informational fairness}\newline
        Providing truthful, understandable, and relevant information about all fairness desiderata across the AI lifecycle.
        & Informational fairness \citep{Colquitt.2001}
        \newline Design publicity \citep{Loi.2021}
        \newline Model cards \citep{Mitchell.2019}
        \newline Outward transparency \citep{Walmsley.2021}\\
    \bottomrule
    \end{tabular}
\end{table}


\subsection{Fairness Understanding}

\begin{quote}
    \small
    \emph{\enquote{Accounting for bias not only requires an understanding of the different sources, that is, data, knowledge bases, and algorithms, but more importantly, it demands the interpretation and description of the meaning, potential side effects, provenance, and context of bias.}}~\citep[p.~8]{Ntoutsi.2020}
\end{quote}

Before mitigating algorithmic unfairness, developers should reflect upon the multidimensional and conflicting nature of fairness, define concrete fairness objectives, and understand how these might be achieved.
In light of this, \emph{fairness understanding} relates to the specification of concrete fairness objectives and knowledge about the socio-technical challenges surrounding the development and deployment of an AI application.

Among others, this may include a sound understanding of fairness schools of thought \citep{Mulligan.2019}, existing social inequalities \citep{Green.2022}, stakeholder-centered fairness requirements \citep{Langer.2021b}, or relevant legal frameworks \citep{Hacker.2022}. 
As the following fairness desiderata will show, there is no one-size-fits-all way in which a system can be fair, both across (e.g., a system can be perceived as fair despite not being formally fair) as well as within fairness desiderata (e.g., formal fairness criteria pose inherent tradeoffs \citep{Friedler.2021}).

\textbf{Stage of the Lifecycle}.
In the AI lifecycle, we map fairness understanding to the initial problem formulation step because all subsequent steps are guided by the fairness objectives discussed and defined in the early stages.
We note, however, that fairness understanding is developed iteratively, based on trials and errors, interdisciplinary exchange, and stakeholder feedback from various development stages---similarly to how a traditional business problem can be better understood and adjusted over the entire course of an AI project.

\textbf{The Role of XAI}. 
XAI can contribute to a better understanding of fairness across all fairness desiderata which, e.g., may lead to a re-evaluation of earlier fairness objectives.
Data-centric explanations \citep{Anik.2021} might reveal pre-existing group disparities (e.g., in the form of differing base rates such as the gender pay gap), which are crucial factors in determining the fairness objectives and how to achieve them.
Further, simulating and testing various formal fairness metrics promotes an understanding of conflicting objectives that may lead to adjustments or re-weightings of fairness objectives \citep{DeArteaga.2022}.
Similarly, XAI may change stakeholders' view of proactively using sensitive features in a specific decision-making context \citep{Nyarko.2021}.
Concerning ``substantive'' fairness \citep{Green.2022, Khan.2022}, XAI may also be applied to normatively gauge the legitimacy of certain features depending on the societal context.

\subsection{Data Fairness}

\begin{quote}
    \small
    \emph{\enquote{Explanations [...] are crucial for helping system developers and ML practitioners to debug ML algorithms by identifying data errors and bias in training data, such as measurement errors and misclassifications, data imbalance, missing data and selection bias, covariate shift, technical biases introduced during data preparation, and poisonous data points injected through adversarial attacks.}}~\citep[p. 248]{Pradhan.2022}
\end{quote}

Undesirable model behavior often stems from flawed data that contains misrepresentations of the world (e.g., erroneous, mislabeled, or imbalanced data) or accurate representations that are societally undesirable (e.g., historical inequalities).
The goal of \emph{data fairness} is to identify and address such flaws in the data set used to train an AI model. 

Prior studies have pointed out data issues as drivers of unfairness \citep{Barocas.2017,Mehrabi.2021,Li.2022} or highlighted data as a starting point for unfairness mitigation \citep{Li.2019,Cai.2020,Saleiro.2020}.
\citet{Mehrabi.2021} provide a comprehensive list of data biases that may introduce unfairness early on in AI development.
Awareness and understanding of these biases are crucial to derive strategies to handle them.

\textbf{Stage of the Lifecycle}.
Data fairness relates to data collection (e.g., in the form of labeling errors, sampling bias, imbalances, etc.) and data analysis, which aims to identify and potentially mitigate unfair data characteristics early on.
However, data fairness also reaches into feature selection which is usually part of an iterative loop together with model construction.
For example, features might be dropped if they are not justifiably task-relevant.
Importantly, in the context of sensitive attributes, developers should always be aware of the flaws of the idea of ``fairness through unawareness''~\citep{Dwork.2012,Nyarko.2021}.

\textbf{The Role of XAI}.
As data is a main source of unfairness \citep{Mehrabi.2021}, XAI is suited to identify potential disparities, imbalances, or abnormalities manifested in the available data early on.
Descriptive statistics are a natural first step to explore pre-existing disparities \citep{Li.2022}.
\citet{Anik.2021} and \citet{Mitchell.2019} provide two examples of how to present simple descriptions and visualizations about the collection, feature distributions, and patterns of a dataset.
XAI techniques have further been claimed to reveal instances and features in the data that have undesirable effects on the model output \citep{Black.2020, Fan.2022, Pradhan.2022} which, however, are often subject to questionable causal and normative assumptions \citep{chiappa2019path, Binns.2020}.
These approaches indicate a strong connection between data fairness and \emph{formal fairness}.

\subsection{Formal Fairness}

\begin{quote}
    \small
    \emph{\enquote{To counteract biases, it is, therefore, crucial to enable their detection. Explainability approaches may aid in this regard by providing means to track down factors that may have contributed to unfair and unethical decision-making processes and either to eliminate such factors, to mitigate them, or at least to be aware of them.}}~\citep[p. 6]{Langer.2021b}
\end{quote}

The most common fairness desideratum is concerned with formal model properties.
By \emph{formal fairness}, we refer to the vast array of formal criteria that have been proposed as mathematical and statistical measures of fairness \citep{Barocas.2019,Castelnovo.2022}.

Consistent with \citet{Verma.2018}, this includes all fairness definitions based on statistical measures (e.g., demographic parity), similarity measures (e.g., fairness through awareness), or based on causal reasoning (e.g., counterfactual fairness).
Formal fairness notions are often distinguished into group and individual fairness.
Group fairness criteria typically require a form of parity between demographic groups, e.g., along sensitive attributes like gender or race~\citep{Chouldechova.2017}.
Individual fairness criteria typically demand to treat similar people alike~\citep{Dwork.2012}.

\textbf{Stage of the Lifecycle}.
Formal fairness is particularly relevant for the iterative loop of model construction and model evaluation.
As soon as the first model prototype is ready, fairness metrics can be evaluated.
Based on the evaluation, unfairness mitigation techniques can be implemented and validated iteratively \citep{chen2023fairness}.

\textbf{The Role of XAI}. 
Regarding formal fairness, XAI could be of exploratory value, offering a plethora of tools and a novel perspective from which to explore formal fairness from multiple angles.
Again, descriptive statistics present a natural first step to identify potential disparities, imbalances, or abnormalities outputted by the model \citep{Chouldechova.2017, Ahn.2020}.
This is especially useful if formal fairness objectives are already specified, e.g., when testing whether formal group fairness metrics are sufficiently satisfied \citep{chen2023fairness}.
XAI might also point towards specific features driving the violation of group fairness metrics \citep{Ghosh.2022} or shed light on more subtle forms of formal fairness such as fairness of recourse \citep{Gupta.2019}.
XAI has further been claimed to elucidate the complex interplay between ``task-relevant'' and correlated ``protected'' features (e.g., \citep{Grabowicz.2022}) which, again, relies on causal and normative assumptions \citep{chiappa2019path, Binns.2020}.
Technical possibilities of XAI to explore formal fairness are manifold but require utmost caution when applied for unfairness mitigation in specific application contexts and should not be misinterpreted as fairness ``proofs'' or ``guarantees''~\citep{Deck.ACritical.2024}.
Particularly, when interpreting the legitimacy of using sensitive information, it is paramount to account for the ``fairness through unawareness'' fallacy~\citep{Dwork.2012, Lipton.2018b} and for differential subgroup validity~\citep{Hunter.1979,Gupta.2023}.

\subsection{Perceived Fairness}

\begin{quote}
    \small
    \emph{\enquote{We argue that only fair systems that are also perceived to be fair by their users should be accepted and employed in practice.}}~\citep[p. 5]{ShulnerTal.2022}
\end{quote}
Another common fairness desideratum refers to positive perceptions of stakeholders, which is often related to trust and acceptance \citep{ShulnerTal.2022, Papenmeier.2019}.
In this sense, \emph{perceived fairness} captures how stakeholders, particularly those affected by a decision, perceive the fairness of the AI system.

Measures of perceived fairness can be derived from the justice constructs of \citet{Colquitt.2001}, which decompose fairness perceptions into a procedural, distributive, and informational dimension (see also \citep{Binns.2018b, Dodge.2019, Schoeffer.2022c}).
Accounting for fairness perceptions promotes a valuable means to design AI systems based on human needs and ideals while giving voice to societal values, which \citet{Rahwan.2018} coined as ``society-in-the-loop''.
Notably, the desirable aspect of perceived fairness does not necessarily or solely lie in positive fairness perceptions but in ``appropriate'' fairness perceptions \citep{Schoeffer.2021}.

\textbf{Stage of the Lifecycle}.
Although fairness perceptions can be relevant at any stage of the lifecycle (e.g., to evaluate data fairness \citep{Anik.2021}), we map fairness perceptions to the model evaluation, deployment, and inference \& usage stage.
Most prior works have measured fairness perceptions after a (mock-up) model has been developed to ask stakeholders about certain outcomes or the AI system itself \citep{Starke.2022}.
Beyond that, there are approaches trying to embed stakeholders' values throughout the conception and development of AI systems (e.g., \citep{Noothigattu.2018, Lee.2019c, Stumpf.2021,Yaghini.2021}).

\textbf{The Role of XAI}.
Explanations may serve stakeholders as a cue to confirm whether their ideas of fairness are implemented in a system.
This idea has also been commonly expressed in prior literature \citep{Ras.2018,ShulnerTal.2022,ShulnerTal.2022b,Papenmeier.2019}.
For example, \citet{Lee.2019b} indicate that explanations can decrease fairness perceptions when they reveal information that stands in conflict with peoples' fairness beliefs.
However, the effect of XAI on fairness perceptions is highly context-dependent and moderated by human factors like political ideology and self-interest \citep{Starke.2022b}.
This indicates that stakeholders require tailored information addressing their case-specific concerns. 
Affected parties, e.g., seem to appreciate information beyond a model's inner workings, such as system context, usage, and data \citep{Schmude.2023}.
We note that optimizing XAI to stimulate positive fairness perceptions isolated from complementary desiderata (e.g., trustworthiness or formal fairness) can lead to undesirable effects such as placebic explanations \citep{eiband2019impact}, fairwashing \citep{Aivodji.2019} or deception \citep{LeMerrer.2020, JohnMathews.2022}.

\subsection{Fairness With Human Oversight}

\begin{quote}
    \small
    \emph{\enquote{Research suggests that neither humans nor ML models are likely to achieve fairness working alone. Instead, human–ML augmentation, where humans and technology work together to perform organizational tasks jointly, is the most promising path to achieving fairness.}}~\citep[p. 2]{Teodorescu.2021}

\end{quote}

Beyond the (formal) fairness of a model itself, fairness can also refer to the decision-making process in which the model is embedded.
\emph{Fairness with human oversight} aims at installing and supporting a human decision-maker to realize case and context-sensitive fairness objectives through human oversight or human discretion.

The human-AI setting can take various forms but usually involves overseeing and overruling unfair outputs or fostering effective reliance behavior \citep{Teodorescu.2021}.
Although legal and ethical guidelines often demand human oversight \citep{EuropeanCommission.HLEG.2019, Enqvist.2023}, the effect of human oversight on fairness is not necessarily beneficial and not well understood yet \citep{Walmsley.2021, Langer.2023, Teodorescu.2021,schoeffer2024explanations}.
Accordingly, this desideratum does not necessarily capture fairness \textit{through} human oversight, but potentially also fairness \textit{despite} human oversight.

\textbf{Stage of the Lifecycle}.
Fairness with human oversight becomes relevant after the model has been deployed, i.e., during inference \& usage.



\textbf{The Role of XAI}.
Where formal implementation of fairness during model development is difficult or human oversight is required, XAI is crucial to inform human discretion so that fairness objectives can still be realized.
In this sense, XAI is commonly proposed to support human decision-makers and domain experts in fostering fairer decisions (e.g., \citep{Stumpf.2021,Slack.2020,Nakao.2022}).
Beyond simple recommendations, such information may include a comparison to similar instances \citep{Chakraborty.2020}, disclosure of uncertainty \citep{Bhatt.2021}, or conditional heatmaps in computer vision tasks \citep{Achtibat.2023}.
To tackle automation bias, \citet{Miller.2023}'s concept of ``evaluative AI'' also highlights providing not only explanations for a certain recommendation but rather balanced evidence for multiple possible outcomes.
However, both designing XAI and training human decision-makers to interact with the explanations is challenging \citep{Lebovitz.2022}.
As feature importance explanations may even hinder fairness of human decisions \citep{schoeffer2024explanations}, we are in need of more conceptual and empirical research on how to design XAI towards fairness objectives in human-in-the-loop settings (e.g., how to effectively override certain types of AI recommendations that violate fairness objectives).

\subsection{Empowering Fairness}

\begin{quote}
    \small
    \emph{\enquote{From the perspective of individuals affected by automated decision-making, we propose three aims for explanations: (1) to inform and help the individual understand why a particular decision was reached, (2) to provide grounds to contest the decision if the outcome is undesired, and (3) to understand what could be changed to receive a desired result in the future, based on the current decision-making model.}}~\citep[p. 2]{Wachter.2017}
\end{quote}

Fairness considerations do not end after an AI model has been designed and decisions have been made.
\emph{Empowering fairness} refers to the ability of affected parties to take effective actions regarding the outcome of a particular decision, e.g., by contesting decisions or seeking recourse.

In her article on the right to explanation, \citet{Vredenburgh.2022} proposes two types of \textit{self-informed advocacy}: retrospectively, affected parties should be able to identify the accountable entity to demand remedy for unfair treatment (viz., responsibility); prospectively, contesting and recourse options should enable affected parties to actively improve upon their possibly unfair outcome (viz., agency).
Our conception of empowering fairness strongly relates to \citeauthor{Vredenburgh.2022}'s forward-looking self-informed advocacy.
It more generally relates to attempts to conceptualize what makes a \emph{fair explanation} in the context of the right to explanation \citep{Hacker.2022}.

\textbf{Stage of the Lifecycle}.
Recourse and contesting is only possible after a decision has been made. Accordingly, we map empowering fairness to inference \& usage.

\textbf{The Role of XAI}.
XAI could be useful for empowering fairness because it may help acquire the information to be communicated.
Contrastive explanations (answering the question ``Why P rather than Q?'') are commonly proposed to provide intuitive entry points to engage with affected parties \citep{Wachter.2017, Miller.2019}.
For instance, counterfactual explanations have been claimed to provide a promising solution to inform and empower affected parties by clarifying what combination of feature values would lead to a different outcome. \cite{Wachter.2017, Asher.2022}.
Therefore, they can provide valuable information required for algorithmic recourse, e.g., by providing actionable recommendations to a loan applicant on what to do to be granted a loan in the future \citep{Karimi.2022, Crupi.Counterfactualexplanations.2022}.
While research on counterfactual explanations is growing rapidly, they do not come without limitations regarding their actionability \citep{Karimi.2022}, validity \citep{Guidotti.2022}, underlying assumptions \citep{Hu.2020}, or susceptibility to manipulations \citep{Slack.2021}.
Extending the scope beyond model-centered explanations, XAI might also involve practical information about responsible contact persons, guidance on how to seek redress, or collaborative platforms for affected parties to share experienced outcomes \citep{LeMerrer.2020}.

\subsection{Long-Term Fairness}

\begin{quote}
    \small
    \emph{\enquote{Accuracy, discrimination, and security characteristics of a system can change over time as well. Simply testing for these problems at training time [...] is not adequate for high-stakes, human-centered, or regulated ML systems. Accuracy, discrimination, and security should be monitored in real-time and over time, as long as a model is deployed.}}~\citep[p. 18]{Gill.2020}
\end{quote}

Fairness remains relevant over the entire AI lifecycle, even after deployment.
Hence, \emph{long-term fairness} captures the dynamic interplay of an AI system with the socio-technical system it is deployed in over time.

The long-term impact of an AI model can be analyzed from several perspectives.
\citet{ArifKhan.2022} contrast ``formal'' and ``substantive'' equality of opportunity where a forward-facing view of fair life chances also accounts for affected parties' future prospects of success (as opposed to ensuring fair contests at a discrete point in time).
Since such conceptions make assumptions about structural disadvantages and future prospects of affected parties, they require much broader and longitudinal evaluation.
Further, due to concept drift \citep{baier2019cope}, a model's formal fairness properties can change over time and should be monitored.

\textbf{Stage of the Lifecycle}.
We map long-term fairness primarily onto inference \& usage, noting that it may encompass all future iterations of the AI lifecycle until the AI system is shut down.

\textbf{The Role of XAI}.
Monitoring tools may help to track changes in fairness metrics and identify situations where interventions are necessary \citep{Castelnovo.2021, Panigutti.2021, Ghosh.2022b}.
Another long-term fairness impact is strategic gaming behavior that arises from transparent models \citep{Lepri.2018}.
For example, loan applicants who receive counterfactual explanations exactly describing how to reach the decision threshold are prone to create a game-theoretic situation where information itself can be unfairly distributed among clients.
We suspect several other forms of long-term fairness issues will emerge that are currently under-explored in the literature.
For example, \citet{Liu.2018b} model the impact of formal fairness on the underlying population over time, which \citet{Hardt.2022} coined as ``performative power''.
Novel forms of XAI may help to anticipate and evaluate such dynamics.

\subsection{Informational Fairness}

\begin{quote}
    \small
    \enquote{\emph{Transparency [...] is valuable because and in so far as it enables the individuals, who are subjected to algorithmic decision-making, to assess whether these decisions are morally and politically justifiable.}}~\citep[p. 254]{Loi.2021}
\end{quote}

All of the listed fairness desiderata can be augmented with a meta desideratum targeted at transparency \emph{about} fairness, which we label as \emph{informational fairness}.

Informational fairness was originally introduced as a psychological construct in the context of organizational justice \citep{Colquitt.2001} to test whether the communication accompanying a decision is candid, truthful, reasonable, timely, and specific.
Our conception of informational fairness is inspired by this construct and corresponds to a great extent to \citet{Loi.2021}'s concept of design publicity.
In line with our idea of broad XAI, \citet{Loi.2021} demand a form of transparency that explains not only a model and its underlying functioning but also the goals and values that went into its design and how these are embedded in the model.
A similar distinction is made by \citet{Walmsley.2021}, who differentiates between functional transparency concerned with the inner workings of an AI model and outward transparency related to communication with stakeholders.

\textbf{Stage of the Lifecycle}.
We conceive informational fairness as a meta desideratum that applies to all other fairness desiderata (see \Cref{fig:fairness_lifecycle}).
Accordingly, informational fairness can be considered across all stages of the AI lifecycle from problem formulation to inference \& usage.

\textbf{The Role of XAI}.
In the case of informational fairness, XAI is not an aid to but the desideratum itself.
Following this idea, we provide examples of what could be made transparent and communicated to affected parties for each fairness desideratum.
Regarding fairness understanding, developers and deployers could justify their fairness objectives, delineate potential tradeoffs, and elucidate the mechanisms through which AI aims to achieve them \citep{Loi.2021}.
For data fairness, they could explain how the training data is composed, highlighting potential sources of bias \citep{Anik.2021}.
Explaining the training objectives and outlining potential uncertainties or goal conflicts might be appropriate to address formal fairness \citep{Mitchell.2019}.
Regarding perceived fairness, explaining the functions and limitations of XAI-generated information could enhance understanding to approach ``appropriate fairness perceptions'' \citep{Schoeffer.2021}.
Transparency about the role of AI and the responsibilities of a human decision-maker within the decision-making context might be desirable for fairness with human oversight \citep{Martin.2019}.
The idea of empowering fairness relies on truthful, understandable, and actionable information \citep{Watson.2021}.
Lastly, regarding long-term fairness, unexpected downstream impacts and the factors driving potentially unfair dynamics can be communicated to stakeholders \citep{Slack.2020}.

\section{The COMPAS Case} \label{sec:compas}


To illustrate the utility of our mapping, we describe how XAI could help address fairness throughout the lifecycle of a high-stakes AI system.
As an example, we consider the recidivism prediction software COMPAS developed by Northpointe (today rebranded to \emph{equivant Supervision}).
While COMPAS is not (only) a traditional AI model, it serves as a fitting example because it is data-driven at least to some extent and a blackbox system due its proprietary status \citep{Rudin.2019}.
COMPAS has been installed in many US-American jurisdictions in order to predict whether a defendant will likely commit another crime in the near future.
Judges use this information, e.g., for decisions about who they release on bail. 
However, COMPAS has been criticized by investigative journalists at ProPublica for disadvantaging Black people \citep{Angwin.2022}.
This is especially problematic given the history of systematic discrimination and marginalization of Black people in the US \cite{pettit.2018}.
In this section, we illustrate both how Northpointe could have addressed different fairness desiderata during system development and how they could still, given COMPAS' continued use, address some of them. 
Of course, XAI is not the only means to this end and is not to be conceived as an ethical panacea.
Furthermore, our illustration also presupposes an inherent motivation to actually address fairness desiderata,  which is not necessarily given in the context of a profit-oriented proprietary system like COMPAS.

Let us proceed along the AI lifecycle.
First, Northpointe could have used XAI to ensure that the development is based on an appropriate \emph{fairness understanding}.
ProPublica's analysis of COMPAS shows that although it was tested for some fairness objective (viz., predictive parity), it falls short on other fairness objectives (viz., equalized odds) in problematic ways \citep{Chouldechova.2017}.
Thus, predictive parity may have been an inadequate fairness objective.
Northpointe could also have based its development on insights about the predictive value and correlations of certain features.
It has been shown that defendants' age and number of previous crimes are most predictive for recidivism \citep{Dressel.2018, Rudin.2019} but also highly correlated to race.
Based on such insights, Northpointe could have taken a position on how to treat such features considering pre-existing systematic discrimination. 

To address \emph{data fairness}, Northpointe would have needed to ensure that the data set adequately represents demographic groups targeted by the system and to be aware of existing structural relationships such as statistically higher crime rates in predominantly Black neighborhoods \citep{Binns.2018}.
Descriptive statistics could have already pointed them towards biases introduced in the data collection process---a reason to reiterate the data collection stage to apply new data collection strategies \citep{Li.2022}.
Statistical analysis could also have helped unveil traces of systematic discrimination in the data (e.g. that Black people are more likely to be arrested for minor offenses due to increased police presence in Black neighborhoods).
Northpointe could have used such insights either to change their data collection strategies (e.g., collecting features that are less correlated to race), or they could have used them during the feature selection phase (e.g., to mitigate systematic discrimination at the data-level \citep{Pradhan.2022}).

Regarding \emph{formal fairness}, Northpointe tested COMPAS for equality of error rates (viz., predictive parity).
By contrast, journalists at ProPublica tested for the balance of true and false positives (viz., equalized odds) \citep{Chouldechova.2017}.
Notably, many formal fairness metrics cannot be satisfied simultaneously \citep{Friedler.2021}, and selecting appropriate formal fairness metrics is an intricate endeavor \citep{DeArteaga.2022}.
However, had Northpointe tested COMPAS for equalized odds, too, they might have anticipated the social backlash and prevented some of the harmful impact.
Today, developers can draw on an extensive suite of fairness testing tools for a comprehensive formal fairness assessment \citep{Linardatos.2020, chen2023fairness, Deck.ACritical.2024}.

Northpointe could have also detected flaws by engaging with stakeholders and gathering feedback on \emph{perceived fairness}.
Outcome statistics, fairness metrics, or model explanations (e.g., in the form of counterfactuals) could have been presented to a focus group of diverse stakeholders to assess and discuss fairness perceptions (see \citep{Stumpf.2021}).
Potential concerns could be handled by reconciling the different perspectives, and even today (after deployment) Northpointe could continue to monitor fairness perceptions by providing feedback channels.


Our discussion thus far has primarily focused on measures that COMPAS' developers could have taken early in the AI lifecycle. 
But even now, during COMPAS' continued usage, Northpointe could still address several fairness issues via XAI. 
First, they could seek \emph{fairness with human oversight}.
This is a particularly relevant desideratum because COMPAS does not make autonomous decisions but informs the decisions of judges.
So, Northpointe could (if not already happening) familiarize judges with the basic functioning, strengths, limitations, and uncertainties of the system as well as the factors driving the risk score to ensure judges will use it appropriately. 
Further, analyzing or supervising the judges' interactions with the system and its explanations might help to achieve fairness objectives more effectively \citep{Teodorescu.2021}.

Northpointe could also foster \emph{empowering fairness} by providing defendants with two kinds of explanations:
First, global explanations \citep{Dodge.2019} about the general functioning of COMPAS could be communicated. Where relevant factors are under a person's control (as, e.g., with defendants' history of misdemeanor or substance abuse), insights about their impact can allow people to make life choices decreasing their risk scores.
Second, local explanations \citep{Dodge.2019} could provide information about an individual's risk score granting the opportunity to (justifiably) contest the score, e.g., by denouncing discriminatory treatment \citep{JohnMathews.2022}, or to (effectively) seek recourse, e.g., by signaling plans to go into rehab \citep{Wachter.2017}. 

Moreover, Northpointe could monitor \emph{long-term fairness} by tracking the specified formal fairness objective(s) over time and accordingly adjust the model over several iterations of the AI lifecycle \citep{Castelnovo.2021}.
Further, changes in defendants' behavior due to specified fairness objectives or increased transparency could be examined.
For example, precise global or counterfactual explanations might allow defendants to game the system in unexpected ways strategically \citep{Lepri.2018}, which might---similar to unfair recourse \citep{Gupta.2019}---be easier for some than for others.

Finally, Northpointe could have aimed and still could aim, to ensure \emph{informational fairness}.
To this end, they would have needed to document and communicate fairness-related information accordingly.
For example, Northpointe could have used model (fairness) cards similar to \citep{Mitchell.2019}, which would have clarified many of the issues that were only revealed due to the investigative work of \citet{Angwin.2022}.
Most importantly, perhaps, Northpointe could have been candid about the underlying business and fairness objectives from the start, ideally making transparent decisions about tradeoffs (e.g., when looking at predictive parity rather than equalized odds) and justifying why an AI system was an appropriate tool for criminal risk prediction in the first place.
In communication with affected parties, COMPAS could be complemented with truthful, understandable, and relevant information \citep{Asher.2022} explaining the underlying logic. 
At the same time, jurisdictions could be more transparent about how exactly COMPAS is employed and how judges interact with the system.
Overall, the strategic use of XAI along the AI lifecycle could benefit all fairness desiderata regarding a model like COMPAS---given sincere intentions to actually pursue these desiderata.

\section{Conclusion and Outlook} \label{sec:conclusion}


We distilled eight fairness desiderata from interdisciplinary literature, mapped them onto the AI lifecycle, and discussed how XAI might be leveraged to contribute to fulfilling them.
Finally, we illustrated the utility of our approach by applying it to the COMPAS case.
The overall picture we paint (see \Cref{fig:fairness_lifecycle}) highlights that to design and utilize XAI for fairness effectively, it is paramount to reflect on which fairness desideratum one seeks to fulfill.
Before closing, we shall briefly comment on some limitations of our proposal and point to avenues for future work.


\textbf{Conceptualization and Validation.}
We do not claim our proposed fairness desiderata to be mutually exclusive or collectively exhaustive.
As our discussion in \Cref{sec:mapping} has shown, different terms elicit different associations in different communities and conceptual overlap between different fairness desiderata seems difficult to avoid altogether (e.g., long-term fairness to some extent includes formal fairness, etc.).
Yet, we are confident that our application to the COMPAS case highlights the general usefulness of our work.
Thus, we firmly believe that our eight fairness desiderata provide a valuable starting point for refinement in follow-up work.
We expect that future interdisciplinary research can---through both conceptual work and validation by application to real-world cases---provide more sharply distinguished categories, capture more fine-grained distinctions, and incorporate an even broader spectrum of perspectives on fairness.
Specific open issues include, e.g., where to locate procedural fairness in our account.
For now, we deliberately excluded that notion as it has inherently differing meanings across (and sometimes even within) disciplines.
Another open question concerns the status of privacy.
Legal scholars might consider data privacy a form of fairness; conceived this way, privacy might qualify as a form data fairness.


\textbf{Generalizability.}
Our discussion has focused primarily on high-risk applications and more traditional decision-support systems.
To what extent does our proposal generalize to low-risk applications? 
And what about generative AI?
Consider Google's multi-modal AI-chatbot Gemini, which has recently received bad publicity for its inaccurate historical depictions---which were driven by misdirected fairness interventions \citep{TheGuardian.2024}.
We suspect that the usage of (seemingly low-risk) AI systems at scale will accumulate to significant societal impacts bearing more subtle threats to fairness such as, e.g., representational harms \citep{Barocas.2017} in the Gemini case.
Thus, although contemporary regulation (like the European AI Act) focuses predominantly on high-risk systems, fairness is an important concern for low-risk applications, too.
Beyond that, the Gemini case illustrates that fairness considerations affect generative AI just as much as more traditional AI systems for decision-making.
Given the rapid advancement and broad adoption of generative AI, we believe that fairness considerations will be indispensable in this context, though they only start to gain traction \citep{Friedrich.2023,neumann2024diverse}.
What makes them particularly challenging is that traditional stakeholder roles may no longer hold: in the Gemini case, users and affected parties are the same people; and ChatGPT users creating personalized chatbots are no longer ``just users'' either. 
Thus, to ensure fairness, future research might not only have to refine fairness desiderata but also rethink stakeholder categories.


\textbf{Actionability.}
We acknowledge that our proposal does not present an actionable process model for fair software engineering; nor does it provide regulatory guidelines to be enforced by legal institutions.
Still, it may serve as a useful roadmap for researchers, developers, and regulators.
For researchers, it provides a starting point for a truly interdisciplinary discourse going beyond discussions of algorithmic fairness focusing on technical aspects of AI systems. 
For developers, our mapping provides urgently needed guidance as to what fairness challenges should be addressed when developing specific guidelines and requirements (see \citet{HackerHalf.2022}).
More specifically, it may help determine under which circumstances human oversight is beneficial \citep{Teodorescu.2021, schoeffer2024explanations} or how responsibilities can be attributed along the AI lifecycle \citep{Raji.2020}.
Naturally, the implications of our approach will vary with the precise settings; so fine-tuning may be needed for different business, societal, and legal contexts.

\bibliography{bibliography}

\end{document}